\documentclass[twoside]{article}

\usepackage[accepted]{aistats2019}
\usepackage{amsthm}
\usepackage{bm}
\usepackage{todonotes}
\usepackage{pgf,tikz}
\usepackage{subcaption}
\usepackage{enumitem}

\usepackage[round]{natbib}

\newtheorem{claim}{Claim}
\newtheorem{theorem}{Theorem}

\newtheorem{proposition}{Proposition}

\theoremstyle{definition}
\newtheorem{definition}{Definition}
\newtheorem{assumption}{Assumption}

\begin{document}

\runningtitle{Multi-Cause Causal Inference: Counterexamples, Impossibility, Alternatives}

\twocolumn[

\aistatstitle{On Multi-Cause Causal Inference with Unobserved Confounding: Counterexamples, Impossibility, and Alternatives}

\aistatsauthor{Alexander D'Amour}

\aistatsaddress{Google AI} ]

\begin{abstract}
Unobserved confounding is a central barrier to drawing causal inferences from observational data.
Several authors have recently proposed that this barrier can be overcome in the case where one attempts to infer the effects of several variables simultaneously. 
In this paper, we present two simple, analytical counterexamples that challenge the general claims that are central to these approaches.
In addition, we show that nonparametric identification is impossible in this setting.
We discuss practical implications, and suggest alternatives to the methods that have been proposed so far in this line of work: using proxy variables and shifting focus to sensitivity analysis.
\end{abstract}

\section{INTRODUCTION}
Estimating causal effects in the presence of unobserved confounding is one of the fundamental challenges of casual inference from observational data, and is known to be infeasible in general \citep{Pearl_Causality_2009}.
This is because, in the presence of unobserved confounding, the observed data distribution is compatible with many potentially contradictory causal explanations, leaving the investigator with no way to distinguish between them on the basis of data.
When this is the case, we say that the causal quantity of interest, or estimand, is not identified.
Conversely, when the causal estimand can be written entirely in terms of observable probability distributions, we say the query is identified.

A recent string of work has suggested that progress can be made with unobserved confounding in the special case where one is estimating the effects of multiple interventions (causes) simultaneously, and these causes are conditionally independent in the observational data given the latent confounder \citep{wang2018blessings,tran2017implicit,ranganath2018multiple}.
The structure of this solution is compelling because it admits model checking, is compatible with modern machine learning methods, models real-world settings where the space of potential interventions is high-dimensional, and leverages this dimensionality to extract causal conclusions.
Unfortunately, this work does not establish general sufficient conditions for identification.

In this paper, we explore some of these gaps, making use of two simple counterexamples.
We focus on the central question of how much information about the unobserved confounder can be recovered from the observed data alone, 
considering settings where progressively more information is available.
In each setting, we show that the information gained about the unobserved confounder is insufficient to pinpoint a single causal conclusion from the observed data. 
In the end, we show that parametric assumptions are necessary to identify causal quantities of interest in this setting.
This suggests caution when drawing causal inferences in this setting, whether one is using flexible modeling and machine learning methods or parametric models.

Despite these negative results, we discuss how it is still possible to make progress in this setting under minor modifications to either the data collection or estimation objective.
We highlight two alternatives.
First, we discuss estimation with proxy variables, which can be used to identify causal estimands without parametric assumptions by adding a small number of variables to the multi-cause setting \citep{Miao_Identifying_2016,louizos2017causal}.
Secondly, we discuss sensitivity analysis, which gives a principled approach to exploring the set of causal conclusions that are compatible with the distribution of observed data.

\section{RELATED WORK}
This paper primarily engages with the young literature on multi-cause causal inference whose primary audience has been the machine learning community. 
This line of work is motivated by several applications, including genome-wide association studies (GWAS) \citep{tran2017implicit}, recommender systems \citep{wang2018deconfounded}, and medicine \citep{ranganath2018multiple}.
\citet{wang2018blessings} include a thorough review of this line of work and application areas.

These papers can be seen as an extension of factor models to causal settings.
Identification in factor models is an old topic.
The foundational results in this area are due to \citet{kruskal1989rank} and were extended to a wide variety of settings by \citet{allman2009identifiability}.
For more elementary results similar to those in our first counterexample, see \citet{bollen1989structural} or similar introductory texts on factor analysis.

The approach taken in this paper is an example of sensitivity analysis, which is a central technique for assessing the robustness of conclusions in causal inference.
One prominent approach, due to \citet{rosenbaum1983assessing} posits the existence of a latent confounder, and maps out the causal conclusions that result when unidentified parameters in this model are assumed to take certain values.
Our second counterexample takes inspiration from the model suggested in this paper.

\section{NOTATION AND PRELIMINARIES}
Consider a problem where one hopes to learn how multiple inputs affect an outcome.
Let $A = (A^{(1)}, \ldots, A^{(m)})$ be a vector of $m$ variables (causes) whose causal effects we wish to infer, and let $Y$ be the scalar outcome variable of interest.
We write the supports of $A$ and $Y$ as $\mathcal A$ and $\mathcal Y$, repsectively.
For example, suppose that $A$ corresponds a set of genes that a scientist could, in principle, knock out by gene editing, where $A^{(k)} = 1$ if the gene remains active and $A^{(k)} = 0$ if it is knocked out.
In this case, the scientist may be interested in predicting a measure of cell growth $Y$ if various interventions were applied.
Formally, we represent this quantity of interest using the $do$-operator:
$$
P(Y \mid do(A)),
$$
which represents the family of distributions of the outcome $Y$ when the causes $A$ are set to arbitrary values in $\mathcal A$ \citep{Pearl_Causality_2009}.

In general, it is difficult to infer $P(Y \mid do(A))$ from observational, or non-experimental, data because there may be background factors, or confounders, that drive both the outcome $Y$ and the observed causes $A$.
We represent these confounders with the variable $U$, with support $\mathcal U$.
In the presence of such confounders, the conditional distribution $P(Y \mid A)$ may be different from the intervention distribution $P(Y \mid do(A))$.
 
If $U$ is observed, the following assumptions are sufficient to identify the intervention distribution:
\begin{itemize}
\item \textbf{Unconfoundedness}: $U$ blocks all backdoor paths between $A$ and $Y$, and
\item \textbf{Positivity}: $P(A=a \mid U) > 0$ almost surely for each $a \in \mathcal A$.
\end{itemize}
Importantly, under these conditions, no parametric assumptions are necessary to identify the intervention distribution. We say that the intervention distribution is nonparametrically identified under these conditions.

The unconfoundedness assumption ensures that the following relation holds between the unobservable intervention distribution and the observable distributions:
\begin{align}
P(Y &\mid do(A = a)) = E[P(Y \mid do(A = a), U)]\notag\\
    &= E[P(Y \mid A = a, U)]\notag\\
    &= \int_{\mathcal U}P(Y \mid A = a, U = u) P(U = u) du\label{eq:causal integral}.
\end{align}
We will call $P(Y \mid A, U)$ the conditional outcome distribution, and $P(U)$ the integrating measure.
Meanwhile, the postivity assumption ensures that all pairs $(A, U)$ are observable in the support of $U$, so that the integrand in \eqref{eq:causal integral} can be evaluated along each point on the path of the integral.
The intervention distribution $P(Y \mid do(A = a))$ is identified under these conditions because \eqref{eq:causal integral} can be written completely in terms of observable distributions.

\section{UNOBSERVED CONFOUNDING AND MULTIPLE CAUSES}

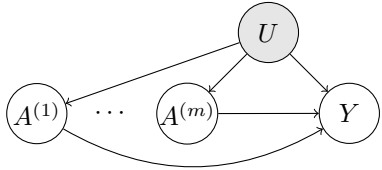
\begin{figure}
\centering
\begin{tikzpicture}[var/.style={draw,circle,inner sep=0pt,minimum size=0.8cm}]
    \node (latent) [var, fill=black!10] {$U$};
    \node (outcome) [var, below right=0.5cm and 0.5cm of latent] {$Y$};
    \node (causem) [var, below left=0.5cm and 0.5cm of latent] {$A^{(m)}$};
    \node (dots) [left of=causem] {$\cdots$};
    \node (cause1) [var, left of=dots] {$A^{(1)}$};
    
    \path[->]
        (latent) edge (outcome)
                 edge (cause1)
                 edge (causem)
        (cause1) edge [bend right] (outcome)
        (causem) edge (outcome);
\end{tikzpicture}
\caption{DAG representation of the multi-cause factor model setting (Assumption~\ref{assn:multicause}).
The causes $A$ obey a factor model, and are conditionally independent given $U$.
In addition, the latent confounder $U$ blocks all backdoor paths between causes $A = (A^{(1)} \cdots A^{(m)})$ and outcome $Y$.
\label{fig:multicause}
}
\end{figure}

When the confounder $U$ is unobserved, the unconfoundedness and positivity assumptions are no longer sufficient for $P(Y \mid do(A = a))$ to be identified.
In this case, additional assumptions are necessary because \eqref{eq:causal integral} is no longer a function of observable distributions.

The multi-cause approach attempts to infer \eqref{eq:causal integral} from the observable data $(A, Y)$ alone under assumptions about the conditional independence structure of this distribution. 
Specifically, this approach incorporates the assumption that the observed distribution of causes $P(A)$ admits a factorization by the unobserved confounder $U$.
We group these central assumptions in Assumption~\ref{assn:multicause} below, and illustrate them in Figure~\ref{fig:multicause}.

\begin{definition}
We say a variable $U$ factorizes the distribution of a set of variables $A$ iff
\begin{equation}
P(A) = \int_{\mathcal U} \left[\prod_{j=1}^m P(A^{(j)} \mid U = u)\right] P(U = u) du \label{eq:factorization},
\end{equation}
\end{definition}

\begin{assumption}
\label{assn:multicause}
There exists an unobserved variable $U$ such that (i) $U$ blocks all backdoor paths from $A$ to $Y$ and (ii) $U$ factorizes the distribution of $A$.
\end{assumption}

Under this assumption, the most general form of multi-cause causal inference rests on the following identification claim.
\begin{claim}
\label{claim:multi cause general}
Under Assumption~\ref{assn:multicause}, for any variable $V$ that factorizes $A$, the following relation $(*)$ holds 
\begin{align}
\int_{\mathcal V} P(Y &\mid A = a, V = v) P(V=v) dv \label{eq:functional}\\
&\stackrel{(*)}{=}\int_{\mathcal U} P(Y\mid A = a, U = u) P(U = u) du\notag\\
&=P(Y \mid do(A = a)).\notag
\end{align}
\end{claim}
If this claim were true, one could obtain an arbitrary factor model for the observed causes satisfying \eqref{eq:factorization} and calculate $P(Y \mid do(A))$.

In Section~\ref{sec:normal example}, we present a simple counterexample that shows that this claim does not hold in general.
The crux of the counterexample is that factorizations of $P(A)$ are not always unique, and differences between these factorizations can induce different values for \eqref{eq:functional}.

In light of this counterexample, it is natural to ask whether identification by \eqref{eq:causal integral} is feasible in the special case that the factorization of $P(A)$ is unique.
In this case, we say the factorization is identified.
Depending on the specification, a factor model may be identified under fairly weak conditions, especially when the latent factor $U$ is categorical; \citet{allman2009identifiability} present a broad set of sufficient conditions. 

\begin{claim}
\label{claim:identified factorization}
Under Assumption~\ref{assn:multicause}, if the factorization of $P(A)$ is identified, then the intervention distribution $P(Y \mid do(A=a))$ is identified by
\eqref{eq:causal integral}.
\end{claim}

\citet{ranganath2018multiple} and \citet{tran2017implicit} make a variation of this claim by supposing that $U$ can be consistently estimated as a function of $A$.
In this case, the factorization is identified in the limit where the number of causes $m$ grows large.

\begin{claim}
\label{claim:consistency}
Under Assumption~\ref{assn:multicause}, if there exists an estimator of $U$ that is a function of $A$, $\hat U(A)$, such that
$$
\hat U(A) \stackrel{a.s.}{\longrightarrow} U,
$$
then the intervention distribution is identified by
\begin{align}
P(Y &\mid do(A = a)) =\notag\\
&\int_{\mathcal U} P(Y \mid A = a, \hat U(A) = u) P(\hat U(A) = u) du.
\label{eq:consistent functional}
\end{align}
\end{claim}

In Section~\ref{sec:impossibility}, we give a counterexample and a theorem showing that Claim~\ref{claim:identified factorization} is false except in the trivial case that the observational and intervention distributions coincide; that is, $P(Y \mid do(A = a)) = P(Y \mid A = a)$.
In a supporting proposition for this result, we show specifically that Claim~\ref{claim:consistency} is false because the consistency premise implies that the positivity assumption is violated.

\section{FACTORIZATION EXISTENCE IS INSUFFICIENT}
\label{sec:normal example}
\subsection{Setup}
In this section, we show that Claim~\ref{claim:multi cause general} is false by a thorough exploration of a counterexample.
Specifically, we show that, even under Assumption 1, it is possible that the observed data is compatible with many distinct intevention distributions $P(Y \mid do(A))$.

Consider a simple setting where all variables are linearly related, and all independent errors are Gaussian.
Letting $\epsilon_{w} \sim N(0, \sigma^2_w)$  for each $w \in \{A, Y, U\}$, the structural equations for this setting are
\begin{align*}
U &:= \epsilon_U\\
A &:= \alpha U + \epsilon_A\\
Y &:= \beta^\top A + \gamma U + \epsilon_Y
\end{align*}
Here, $\alpha, \beta$ are $m \times 1$ column vectors, and $\gamma$ is a scalar;
$\epsilon_A$ is a $m \times 1$ random column vector, and $\epsilon_Y, \epsilon_U$ are random scalars.
This data-generating process satisfies Assumption~\ref{assn:multicause}.

Under this model, the intervention distribution has the following form:
$$
P(Y \mid do(A = a)) = N(\beta^\top a, \gamma^2 \sigma^2_U + \sigma^2_Y).
$$
We will focus specifically on estimating the conditional $E[Y \mid do(A = a)]$, which is fully parameterized by $\beta$.
Thus, our goal is to recover $\beta$ from the distribution of observed data.

The covariance matrix can be written as
$$
\Sigma_{AYU} = \begin{pmatrix} 
\Sigma_{UU} &\Sigma_{UA} &\Sigma_{UY}\\
\Sigma_{AU} &\Sigma_{AA} &\Sigma_{AY}\\
\Sigma_{YU} &\Sigma_{YA} &\Sigma_{YY}
\end{pmatrix}
$$
where $\Sigma_{AA}$ is $m \times m$, $\Sigma_{AY} = \Sigma_{YA}^\top$ is $m \times 1$, and $\Sigma_{YY}$ is $1 \times 1$.

The marginal covariance matrix of the observable variables $(A,Y)$ is the bottom-right $2 \times 2$ sub-matrix of this matrix.
Its entries are defined by:
\begin{align*}
\Sigma_{AA} &= \alpha \alpha^\top \sigma^2_U + \operatorname{diag}(\sigma^2_A)\\
\Sigma_{AY} &= \Sigma_{AA}\beta + \gamma \sigma^2_U \alpha\\
\Sigma_{YY} &= (\beta^\top \alpha + \gamma)^2 \sigma^2_U+ \beta^\top \operatorname{diag}(\sigma^2_A)\beta + \sigma^2_Y
\end{align*}
In these equations, the quantity on the LHS is observable, while the structural parameters on the RHS are unobservable.
The goal is to invert these equations to obtain a unique value for $\beta$.

\subsection{Equivalence Class Construction}

When $m \geq 3$, the number of equations in this model exceeds the number of unknowns, but there still exists an equivalence class of structural equations with parameters
$$(\alpha_1, \beta_1, \gamma_1, \sigma^2_{U,1}, \sigma^2_{A,1}, \sigma^2_{Y,1})
\neq (\alpha, \beta, \gamma, \sigma^2_U, \sigma^2_A, \sigma^2_Y)$$
that induce the same observable covariance matrix, and for which $\beta_1 \neq \beta$.
These parmeterizations cannot be distinguished by observed data.
In this section, we show how to construct such a class.

The key to this argument is that the scale of $U$ is not identified given $A$, regardless of the number of causes $m$.
This is a well-known non-identification result in confirmatory factor analysis  \citep[e.g.,][Chapter 7]{bollen1989structural}.
In our example, the expression for $\Sigma_{AA}$ does not change when $\alpha$ and $\sigma^2_U$ are replaced with the structural parameters $\alpha_1$ and $\sigma^2_{U,1}$:
\begin{align*}
\alpha_1 &:= c \cdot \alpha\\
\sigma^2_{U,1} &:= \sigma^2_{U} / c^2.
\end{align*}
In the following proposition, we state how the remaining structural variables can be adjusted to maintain the same observable covariance matrix when the scale $c$ is changed. 

\begin{proposition}
For any fixed vector of parameters $(\alpha, \beta, \gamma, \sigma^2_U, \sigma^2_A, \sigma^2_Y)$ and a valid scaling factor $c$ (defined below), there exists a vector of parameters that induces the same observable data distribution.
\begin{align}
\alpha_1(c) &= c \cdot \alpha\notag\\
\beta_1(c) &= \beta + \Sigma_{AA}^{-1}\alpha \cdot \gamma\sigma^2_U\left(1-\frac{1}{c}\right)
\label{eq:beta_shift}\\
\gamma_1(c) &= \gamma\notag\\
\sigma^2_{U,1}(c) &= \sigma^2_U / c^2\notag\\
\sigma^2_{A,1}(c) &= \sigma^2_A\notag\\
\sigma^2_{Y,1}(c) &= \Sigma_{YY} - (\beta_1^\top \alpha_1 + \gamma_1)^2\sigma^2_{U,1}\notag\\
 &\quad- \beta^\top_1\operatorname{diag}(\sigma^2_{A,1})\beta_1\notag
 \end{align}
 The factor $c$ is valid if it implies positive $\sigma^2_{Y,1}(c)$.
\end{proposition}

We call the set of all parameter vectors that correspond to valid values of $c$ the \emph{ignorance region} in the parameter space.
Parameters in the ignorance region cannot be distinguished on the basis of observed data because they all imply the same observed data distribution.

\begin{figure}
\centering
\includegraphics[width=1.05\columnwidth]{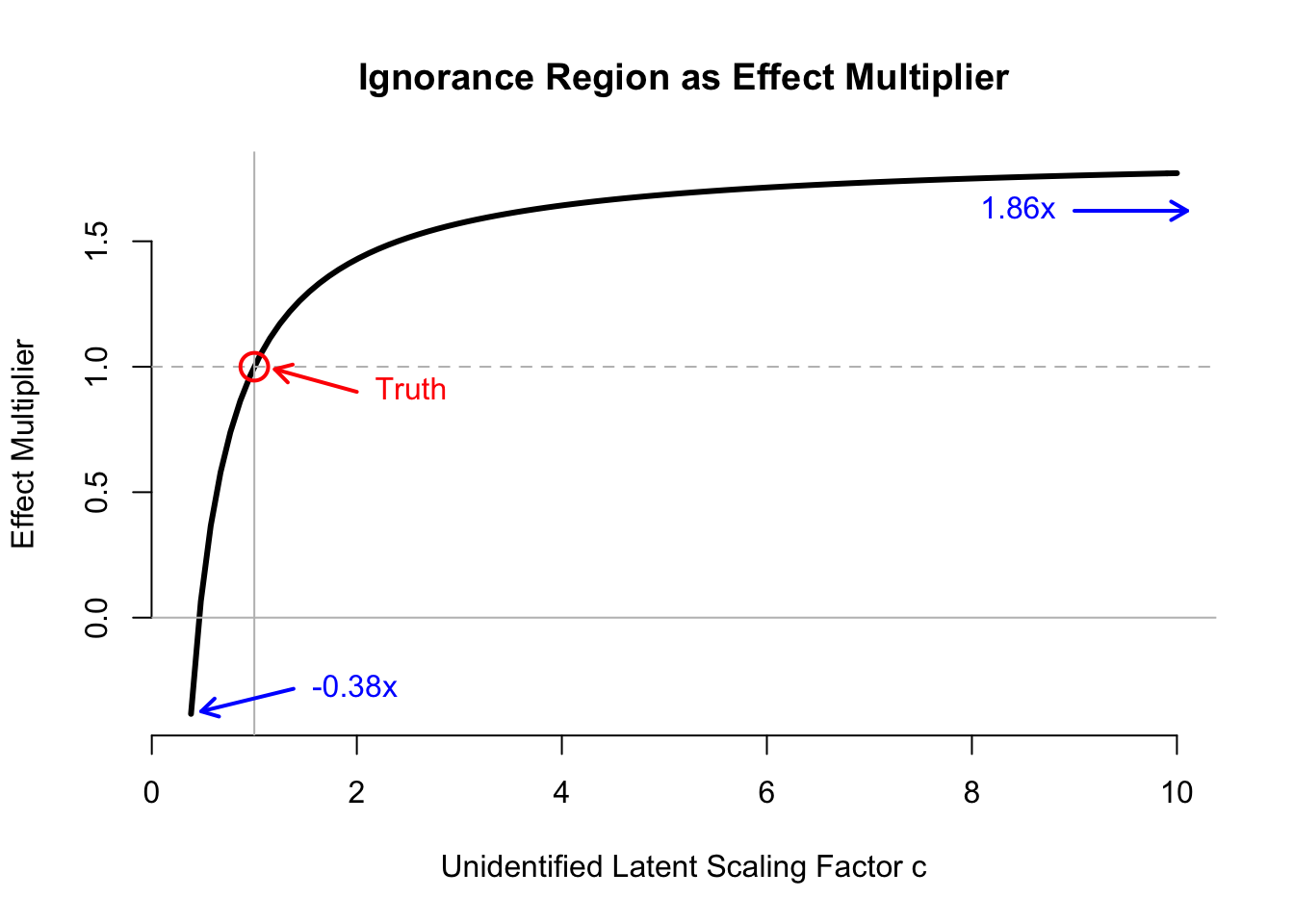}
\caption{%
An illustration of the ignorance region for $\beta$ in an example where the ignorance region can be represented as a scalar effect multiplier.
All values of $\beta$ on the black line are equally supported by the data, and include both positive and negative effects of varying magnitudes.
\label{fig:ignorance}
}
\end{figure}

We plot an illustration of the ignorance region for $\beta$ from a numerical example in Figure~\ref{fig:ignorance}.
In this example, we set $\beta = b \cdot \bm{1}_{m\times1}$ and $\alpha = a \cdot \bm{1}_{m\times1}$ to be constant vectors for some constants $a$ and $b$.
In this case $\beta_1(c) = s(c) \cdot b \cdot \bm {1}_{m\times1}$ is a simple scaling of $\beta$, so the ignorance region can be represented by the value of this scalar $s(c)$.
In this example, the data cannot distinguish between effect vectors that have the opposite sign of the true effect vector $\beta$, and those those that overstate the effect of $A$ by nearly a factor of 2.

\subsection{Large-$m$ Asymptotics}
\label{sec:asymptotics}

The ignorance region does not in general disappear in the large treatment number (large-$m$) limit.
Here, we extend our example to an asymptotic frame where the ignorance region maintains the same (multiplicative) size even as $m$ goes to infinity.
Consider a sequence of problems where the number of treatments analyzed in each problem is increasing in the sequence.
Each problem has its own data generating process, with some structural parameters indexed by $m$: $(\alpha_m, \beta_m, \gamma, \sigma^2_U, \sigma^2_{A,m}, \sigma^2_Y)$.
We keep the scalar parameters not indexed by $m$ fixed.

We consider the marginal variance of each $A$ to be fixed, so for some fixed scalar $s^2_0$, for each problem $m$,
$$
\sigma^2_{A,m} =\bm{1}_{m\times 1} s^2_0.
$$

Likewise, we expect the marginal variance of $Y$ to be relatively stable, no matter how many treatments we choose to analyze.
Given our setup, this means that if the number of treatments is large, the effect of each individual treatment on average needs to become smaller as $m$ grows large, or else the variance of $Y$ would increase in $m$ (this is clear from the specification of $\Sigma_{YY}$).
To handle this, we fix some constant scalars $a_0$ and $b_0$ and assume that, for problem $m$,
$$
\alpha_m = \bm{1}_{m\times 1} \cdot a_0 / \sqrt{m};\quad 
\beta_m = \bm{1}_{m\times 1} \cdot b_0 / \sqrt{m}.
$$
Thus, as $m \rightarrow \infty$, the norms of $\alpha_m$ and $\beta_m$, as well as their inner product $\alpha_m^\top \beta_m$, which appears in the expression for $\Sigma_{YY}$, remain fixed.
\footnote{
The asymptotic frame in this section is not the only way to maintain stable variance in $Y$ as $m$ increases.
In particular, one could specify the sequence of problems so that they are projective, and simulate an investigator incrementally adding causes to a fixed analysis.
One could then define a sequence of coefficients $\alpha^{(k)}$ for each cause added to the analysis, putting some conditions on the growth of the inner product $\alpha_m^\top \beta_m$ and norm of $\beta_m$, such as a sparsity constraint on $\beta_m$.
Our setup here is simpler.
}

Under this setup, the interval of valid values for the latent scaling factor $c$ remains fixed for any value of $m$.%
For a fixed $c$ in this interval, we examine how the corresponding shift vector $\Delta_{\beta, m}(c) = \beta_1(c) - \beta$ behaves as $m$ grows large.
The components of the shift $\Delta_{\beta, m}(c)$ scale as $m^{-1/2}$.
Specifically, applying the Sherman-Morrison formula,
\begin{align*}
\Delta_{\beta,m}(c) &= \Sigma_{AA}^{-1}\alpha_m \cdot \gamma\sigma^2_U\left(1-\frac{1}{c}\right)\\
    &= m^{-1/2} \cdot\bm{1}_{m\times 1} \cdot \frac{a_0}{s^2_0+\sigma^2_U a_0^2}  \cdot \gamma \sigma^2_U\left(1 - \frac{1}{c}\right).
\end{align*}
Thus, for each $k$, the ratio of the $k$th component of the shift vector relative to the $k$th component of the true parameters remains fixed in $m$:
$$
\frac{\Delta_{\beta, m}^{(k)}(c)}{\beta_m^{(k)}} = \frac{a_0}{b_0(s^2_0+\sigma^2_U a_0^2)}  \cdot \gamma \sigma^2_U\left(1 - \frac{1}{c}\right).
$$

Thus, even asymptotically as $m \rightarrow \infty$ there is no identification.

\section{FACTORIZATION UNIQUENESS IS INSUFFICIENT}
\label{sec:impossibility}
\subsection{Impossibility of Nonparametric Identification}
In this section, we consider identification in the multi-cause setting in the special case where the factorization of $P(A)$ by $U$ is unique;
that is, we consider the case where $P(A)$ can be decomposed uniquely into a mixing measure $P(U)$ and a conditional treatment distribution $P(A \mid U)$.
In this setting, Claims~\ref{claim:identified factorization} and \ref{claim:consistency} assert that $P(Y \mid do(A))$ is identified by \eqref{eq:causal integral}.
This claim arises as a natural response to the counterexample in the last section, where the non-uniqueness of the factorization contributes to non-identification.

As in the last section, we show via counterexample that the conditions in these claims are insufficient for identification of $P(Y \mid do(A))$.
In addition, we show that, in general, parametric assumptions about the conditional outcome distribution $P(Y \mid U, A)$ are necessary to identify $P(Y \mid do(A))$, except in the case where there is no confounding, i.e., the intervention distribution $P(Y \mid do(A))$ and the observed conditional distribution $P(Y \mid A)$ are equal almost everywhere.
We summarize this statement in the following theorem.

\begin{theorem}
\label{thm:impossible}
Suppose that Assumption~\ref{assn:multicause} holds, that $P(U, A)$ is identified, and that the model for $P(Y \mid U, A)$ is not subject to parametric restrictions.

Then either $P(Y \mid do(A)) = P(Y \mid A)$ almost everywhere, or $P(Y \mid do(A))$ is not identified.
\end{theorem}

Put another way, our theorem states that nonparametric identification is impossible in this setting, except in the trivial case.
In this section, we prove two supporting propositions for this theorem and demonstrate them in the context of our running example.
The proof of the theorem, which follows almost immediately from these propositions, appears at the end of the section.

\subsection{Counterexample Setup}
Let $U$ be a binary latent variable, and $A := (A^{(1)}, \cdots, A^{(m)})$ a vector of $m$ binary causes.
In the structural model, we assume that the individual causes $A^{(k)}$ are generated independently and identically as a function of $U$.
Let the outcome $Y$ be binary, and generated as a function of $U$ and $A$.
\begin{align*}
U &:= \mathrm{Bern}(\pi_U)\\
A^{(k)} &:= \mathrm{Bern}(p_A(U)) \quad k = 1,\cdots,m\\
Y &:=\mathrm{Bern}(p_Y(U, A))
\end{align*}
In addition to this structural model, we assume that $m \geq 3$ and that $p_A(U)$ is a non-trivial function of $U$.
These assumptions are sufficient for the factorization of $P(A)$ by $U$ to be unique \citep{kruskal1989rank,allman2009identifiability}.
Thus, this example satisfies the premise of Claim~\ref{claim:identified factorization}.

Our goal is to estimate the intervention distribution for each $a \in \mathcal A$ using the identity in \eqref{eq:causal integral}.
Here, the intervention distribution can be summarized by the following causal parameter:
\begin{align}
\pi_{Y \mid do(a)} &:= P(Y=1 \mid do(A=a))\notag\\
&=(1-\pi_U) p_Y(0, a) + \pi_U p_Y(1, a).\label{eq:binary causal integral}
\end{align}
Because the factorization of $P(A)$ is identified, $\pi_U$ and $p_A(U)$ are identifiable.
Thus, to calculate \eqref{eq:binary causal integral}, it remains to recover $p_Y(U, A) = P(Y = 1 \mid U, A)$.

We will show that this conditional probability cannot be recovered from the observed data.
Our approach will be to analyze the residual distribution $P(Y, U \mid A=a)$.
For each value $A = a$, we can characterize $P(U, Y \mid A=a)$ by 4 probabilities in a $2 \times 2$ table (see Figure~\ref{fig:2by2}).
We use shorthand notation $p_{uy \mid a} := P(U = u, Y = y \mid A = a)$ to denote the value in each cell of this table.
The values in this table are unobservable, but they are subject to several constraints.
The entries of this table are constrained to be positive and sum to 1.
In addition, because $P(Y, A)$ and $P(U, A)$ are identified, the margins of the table are constrained to be equal to probabilities given by $P(Y \mid A = a)$ and $P(U \mid A = a)$.
We refer to these probabilities with the shorthand $\pi_{Y \mid a} := P(Y = 1\mid A  = a)$ and $\pi_{U \mid a} := P(U = 1\mid A = a)$.
Using this notation, we can rewrite \eqref{eq:causal integral} as:
\begin{align}
\pi_{Y \mid do(a)} =
(1-\pi_U) \frac{p_{01\mid a}}{1 - \pi_{U \mid a}} + \pi_U \frac{p_{11 \mid a}}{\pi_{U \mid a}}. \label{eq:binary causal shorthand}
\end{align}

We consider identification in two separate cases, depending on the amount of information about $U$ is contained in the event $A=a$.
We first consider the case where $P(U \mid A=a)$ is non-degenerate, so that there is residual uncertainty about $U$ after $A$ is observed.
We then consider the degenerate case, where $U$ can be deterministically reconstructed as a function of $A$ (the premise of Claim~\ref{claim:consistency}).

\begin{figure}
\centering
\begin{tikzpicture}[box/.style={draw,rectangle,text width=1.5cm, minimum height=1.5cm, align=left},
nobox/.style={text width=1.5cm, minimum height=1.5cm, align=center, fill=black!10}]
\matrix (conmat) [row sep=-0.005cm,column sep=-0.005cm, ampersand replacement = \&] {
\node (p00) [box, align = center,
    label=left:{$U=0$},
    label=above:{$Y=0$}
    ] {$p_{00 \mid a}$};
\&
\node (p01) [box, align = center,
    label=above:{$Y=1$},
        ] {$p_{01 \mid a}$};
\& \node(pinotU) [nobox, label={[shift={(0.17,0)}]above:{\footnotesize $P(U \mid A=a)$}}] {$1 - \pi_{U \mid a}$};
\\
\node (p10) [box, align = center,
    label=left:{$U=1$},
        ] {$p_{10 \mid a}$};
\&
\node (p11) [box, align = center,
        ] {$p_{11 \mid a}$};
\& \node(piU) [nobox] {$\pi_{U \mid a}$};
\\
\node (pinotY) [nobox, label=left:{\footnotesize $P(Y \mid A=a)$}] {$1 - \pi_{Y \mid a}$};
\& \node (piY) [nobox]{$\pi_{Y \mid a}$};
\&
\\
};
\end{tikzpicture}
\caption{$2 \times 2$ table representation of $P(U, Y \mid A)$, the joint distribution of confounder $U$ and outcome $Y$ conditional on causes $A$, in the all-binary example.
When the factor model for $P(U)$ is identified, the observed data fix the margins of the table, but leave one degree of freedom unidentified.
\label{fig:2by2}
}
\end{figure}

\subsection{Non-Degenerate Case}
When $P(U \mid A = a)$ is not degenerate, the table in Figure~\ref{fig:2by2} is underdetermined by the data, and has one remaining degree of freedom.
This degree of freedom determines the dependence between $Y$ and $U$ conditional on $A$.
We parameterize this dependence in terms of $p_{11 \mid a} := P(U = 1, Y = 1 \mid A = a)$.
For purposes of interpretation, $p_{11 \mid a}$ is linearly related to the $cor(U, Y, \mid A = a)$ \citep[Sec 7.1.1]{joe1997multivariate}:
$$
cor(U, Y \mid A = a) = \frac{p_{11 \mid a} - \pi_{U \mid a}\pi_{Y \mid a}}{\pi_{U \mid a}(1-\pi_{U \mid a})\pi_{Y \mid a}(1-\pi_{Y \mid a})}.
$$
The constraints on the table constrain $p_{11 \mid a}$ to lie in the following range \citep{joe1997multivariate}:
\begin{align}
\max\{0, \pi_{U \mid a} + \pi_{Y \mid a} - 1\} \leq p_{11 \mid a} \leq \min\{\pi_{U \mid a}, \pi_{Y \mid a}\}.\label{eq:p11 constraint}
\end{align}
Fixing a value for $p_{11 \mid a}$ in this range determines the values of all four cells in the table which, in turn, determine the causal parameter $\pi_{Y \mid do(a)}$ by \eqref{eq:binary causal shorthand}.

\begin{figure*}
\begin{subfigure}[b]{0.595\textwidth}
\centering
\includegraphics[width=1\textwidth]{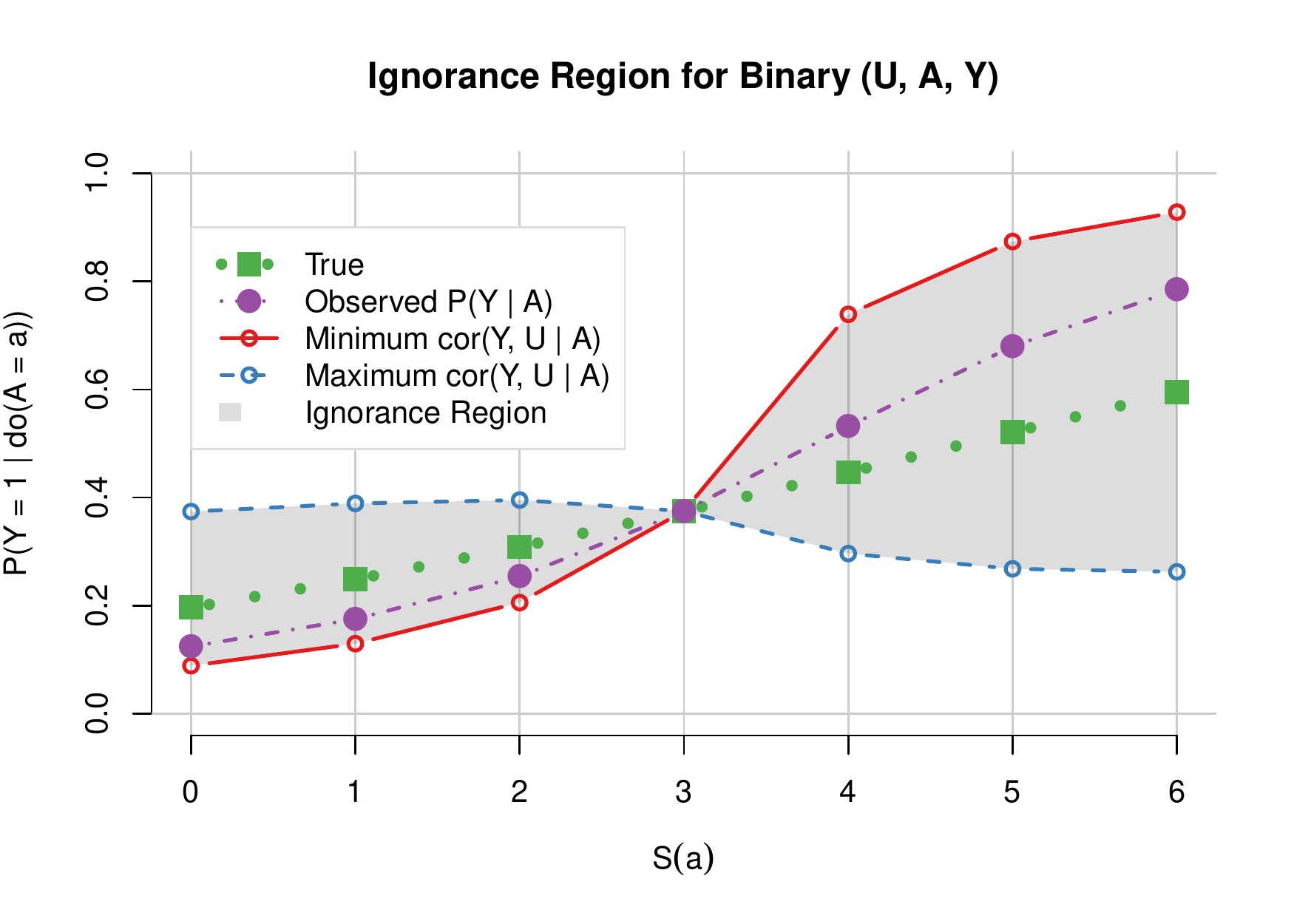}
\end{subfigure}
\begin{subfigure}[b]{0.425\textwidth}
\centering
\includegraphics[width=1\textwidth]{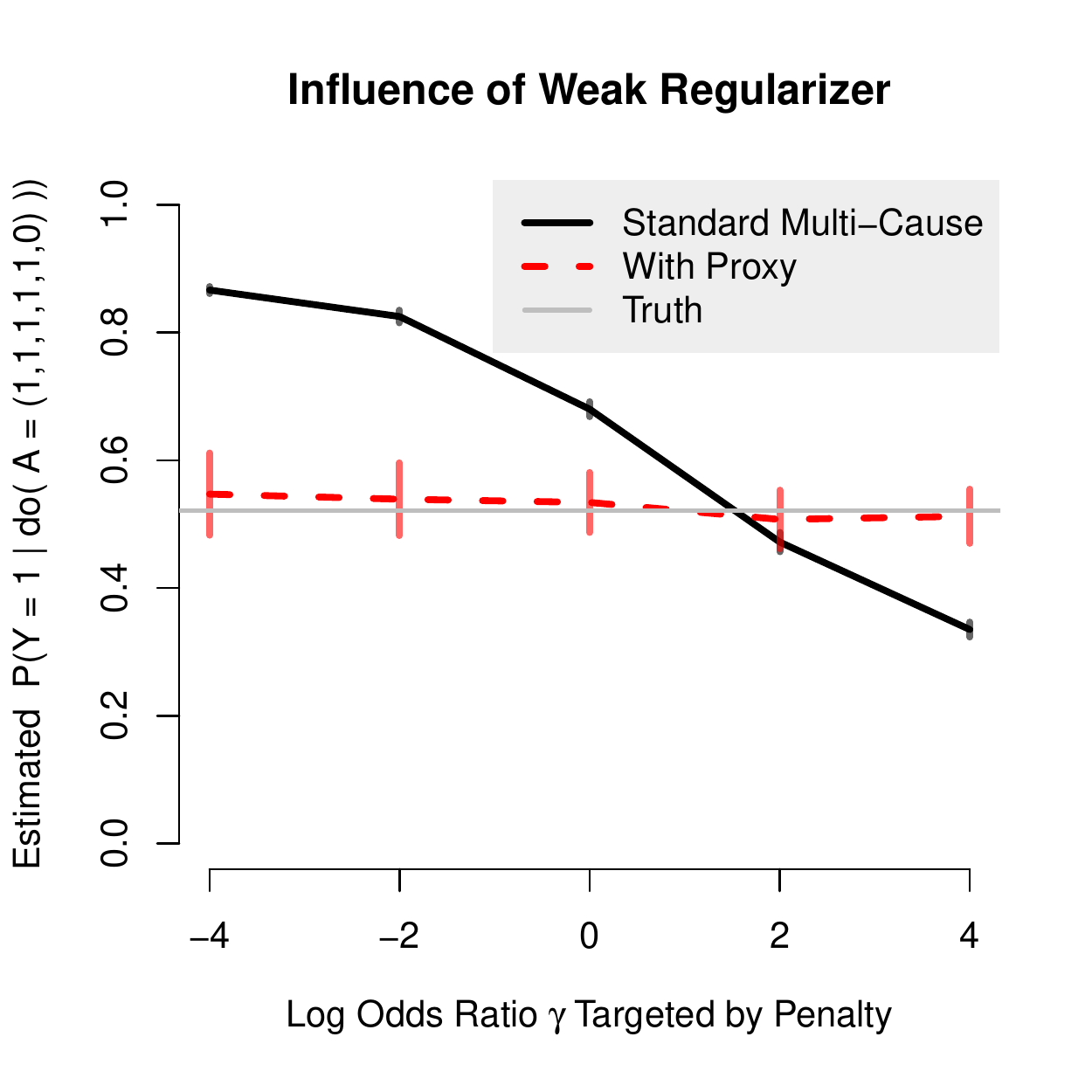}
\end{subfigure}
\caption{(Left) Ignorance regions corresponding to the causal parameters $\pi_{Y \mid do(a)}$, indexed by $S(a) = \sum_{k=1}^m a^{(k)}$. For convenience, this example was instantiated so that these ignorance regions only depend on $S(a)$. The ignorance regions are generated by allowing the unidentified parameter $p_{11 \mid a}$ to vary in its valid range.
(Right) Results of estimation experiment, attempting to recover $P(Y \mid do(A = (1, 1, 1, 1, 1, 0))$ using maximum likelihood estimation. In the standard multi-cause setting, a weak regularizer on $\gamma := \log\left(\frac{p_{11 \mid a}}{p_{10 \mid a}}\left / \frac{p_{01 \mid a}}{p_{00 \mid a}}\right.\right)$ can push the estimate of $\pi_{Y \mid do(a)}$ to arbitrary locations in the ignorance region. When two proxy variables are added, the weak regularizer has little effect. Vertical bars show $\pm$1 sd from 20 simulations.
\label{fig:binary_ignorance}}
\end{figure*}

By varying $p_{11 \mid a}$ in the valid range \eqref{eq:p11 constraint}, we can generate the ignorance region of $\pi_{Y \mid do(a)}$ values that are equivalently compatible with the observed data.
To demonstrate, we instantiate the model with $m=6$ causes.
To simplify plotting, we specify $P(Y \mid U, A)$ so that it only depends on $S(A) = \sum_{k=1}^m A^{(k)}$.
This ensures that inferences about $\pi_{Y \mid do(a)}$ only depend on $S(a)$.
In Figure~\ref{fig:binary_ignorance}, we plot ignorance regions for $\pi_{Y \mid do(a)}$ indexed by $S(a)$.

The figure demonstrates that in this simple example, $P(Y \mid do(A))$ is not identified in general.
Specifically, for all vectors $a$ where $S(a) \neq 3$, the ignorance region is non-trivial.
For each $S(a)$, the extreme points of the ignorance region correspond to the extreme values of $p_{11\mid a}$ compatible with the margins given by $P(Y \mid A)$ and $P(U \mid A)$.
The true causal parameters $P(Y = 1 \mid do(A = a))$ and the observational parameters $P(Y = 1 \mid A = a)$ are always contained in this region.
In some cases, the ignorance region is large relative to the $[0, 1]$ interval.

Figure~\ref{fig:binary_ignorance} also shows a trivial case, where $P(Y \mid do(A = a)) = P(Y \mid A = a)$, and thus the intervention distribution is identified despite $p_{11\mid a}$ being underdetermined.
This case arises because we defined the structural parameters such that $p_A(1) = 1 -p_A(0)$, which ensures that when $S(a) = m/2 = 3$, $P(U \mid A=a) = P(U)$, and thus $P(Y \mid do(A=a)) = P(Y \mid A=a)$.

\subsection{Copula Non-Identification in General}
The non-identification in the above example occurs generally when $P(Y \mid U, A)$ is nonparametric.
We state this in the following supporting proposition for Theorem~\ref{thm:impossible}.
For the general case, we represent the underdetermined degree of freedom by the copula density:
$$
c(Y, U \mid A) := \frac{P(Y, U \mid A)}{P(Y \mid A)P(U \mid A)},
$$
which specifies the dependence between $Y$ and $U$ given $A$, independently of the margins $P(Y \mid A)$ and $P(U \mid A)$ \citep[see, e.g.,][]{nelsen2007introduction}.
In the following proposition, we show in a very general setting that this copula is not identified, and that this non-identification precludes identification of $P(Y \mid do(A))$.

\begin{proposition}
\label{prop:copula}
In the setting of Theorem~\ref{thm:impossible}, suppose that $P(U \mid A)$ is almost surely non-degenerate. Then, the following are true
\begin{enumerate}
\item The copula density $c(Y, U \mid A)$ is not identified.
\item Either $P(Y \mid do(A)) = P(Y \mid A)$, or $P(Y \mid do(A))$ is not identified.
\end{enumerate}
\begin{proof}
For the first statement, the joint distribution $P(U, Y, A)$ can be written
$$
P(U, A, Y) = P(A) P(Y \mid A) P(U \mid A) c(Y, U \mid A),
$$
By assumption, $P(Y, A)$ and $P(U, A)$ are identified, but the copula density $c(Y, U \mid A)$ remains unspecified because there are no restrictions on $P(Y \mid U, A)$.

For the second statement, note that the independence copula $c(Y, U \mid A) = 1$ is compatible with the observed data, as a result of the first statement.
Under the independence copula, $P(Y \mid do(A)) = P(Y \mid A)$.
If this causal hypothesis is not true, then the true $P(Y \mid do(A))$ is also compatible with the observed data, so multiple causal hypotheses are compatible with the observed data, and $P(Y \mid do(A = a))$ is not identified.
\end{proof}
\end{proposition}

\subsection{Degenerate Case}
We now continue our example and consider the case, invoked in Claim~\ref{claim:consistency}, where $P(U \mid A)$ is degenerate almost everywhere, i.e., where for any observable unit, $U$ can be written as a deterministic function of $A$.
In this case, the copula $P(U, Y \mid A)$ is trivial, so the non-identification in Proposition~\ref{prop:copula} is not an issue.
However, $P(Y \mid do(A))$ remains unidentified in this setting because the degeneracy of $P(U \mid A)$ implies that the positivity assumption is violated. 

This case is invoked asymptotically, so we analyze our example as the number of causes grows large.
First, we show that $U$ can indeed be consistently estimated from $A$.
As $m$ grows large, the unobserved confounder $U$ can be estimated perfectly for any unit, up to label-switching.
Specifically, by the strong law of large numbers, for any unit with $U = u$, as $m$ grows large,
$$
\hat p(A) := S(A) / m \stackrel{a.s.}{\rightarrow} p_A(u).
$$
From this fact, and our assumption that $p_A(u)$ is a non-trivial function of $u$, we can construct consistent estimators $\hat U(A)$ for $U$.
For example, letting $I\{\cdot\}$ be an indicator function,
$$
\hat U(A) := I\left\{\hat p(A) > \frac{p_A(1) + p_A(0)}{2}\right\}
$$
is consistent as $m$ grows large.

\begin{figure*}
\includegraphics[width=\textwidth]{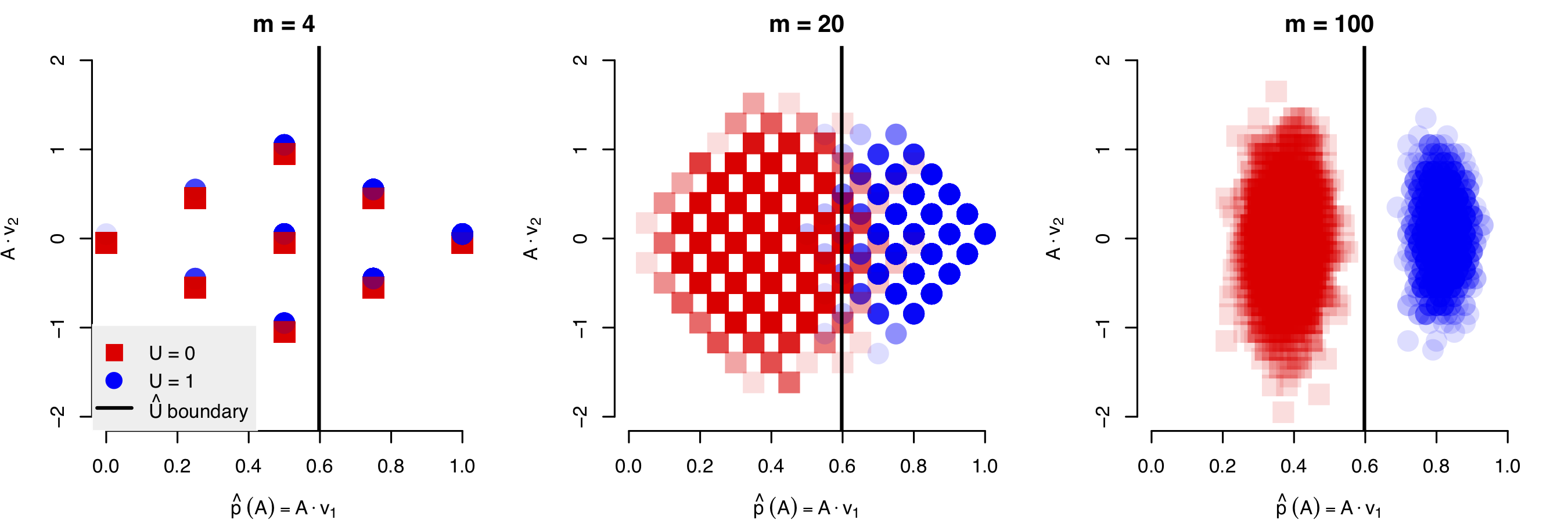}
\caption{Linear projections of sampled treatments $A$, with color and shape indicating the value of the latent confounder $U$.
Here, $v_1 = m^{-1} \cdot \bm{1}_{m\times 1}$, such that $A^\top v_1 = S(A)/m = \hat p(A)$, and $v_2$ is a vector orthogonal to $v_1$, scaled by $m^{-1/2}$.
These values are generated according to the model in Section~\ref{sec:impossibility}.
As $m$, the dimension of $A$, increases, the treatment vectors $A$ generated under each value of $U$ become more distinct.
We take advantage of this distinctness to estimate $U$ consistently; as $m$ grows large, the black boundary encodes an estimator $\hat U(A)$ that calls $U=0$ if a point lands to the left of the line, and $U=1$ if it lands to the right.
However, this separation between red squares and blue circles also indicates a positivity violation.
\label{fig:binary positivity}
}
\end{figure*}

However, as $m$ grows large, the causes $A$ begin to occupy disjoint regions of the cause space $\mathcal A$, inducing a violation of the positivity assumption.
This is the same phenomenon that drives the consistency of $\hat U(A)$.
We illustrate in Figure~\ref{fig:binary positivity}, where we plot samples of the causes $A$ at various values of $m$, projected into two dimensions and scaled. 
The first dimension is obtained by calculating $\hat p(A)$, expressed as a linear operator: $\hat p(A) = A^\top (m^{-1} \cdot \bm 1_{m \times 1})$.
The second dimension is obtained by projecting $A$ onto a vector orthogonal to $\bm 1_{m \times 1}$.
In this case, we choose the vector $v_2 = m^{-1/2} \cdot (\bm 1_{(m/2) \times 1}^\top, -\bm 1_{(m/2) \times 1}^{\top})^\top$ and calculate $A^\top v_2$.
When $m$ is small, $A$ has the same support whether $U = 0$ or $U = 1$, but as $m$ grows, the causes $A$ concentrate on either side of the decision boundary of our estimator $\hat U(A)$, ensuring consistency of the estimator, but violating positivity.

Because of this phenomenon, the conditional probabilities
$P(Y=1 \mid U=0, \hat U(A)=1)$ and $P(Y=1 \mid \hat U(A)=0, U=1)$ are inestimable from the data in the large-$m$ limit, and we cannot evaluate \eqref{eq:binary causal integral}. 
At best, we can bound \eqref{eq:binary causal integral} by substituting extreme values $0$ and $1$ for the inestimable probabilities.

This problem is not merely asymptotic; it also manifests for finite $m$, when the residual uncertainty about $U$ after observing $A$ is small.
Consider the $2 \times 2$ table in Figure~\ref{fig:2by2} corresponding to a cause vector $a$ for which $\pi_{U \mid a} = \epsilon$ for some small epsilon satisfying $\epsilon < \min\{\pi_{Y \mid a}, 1-\pi_{Y \mid a}\}$.
The bounds in \eqref{eq:p11 constraint} imply that $p_{11 \mid a} \in [0, \epsilon]$, such that
$$P(Y = 1 \mid U = 1, A = a) \in [0, 1],$$
i.e., the data provide no information about this conditional probability.
In the limit, as $\epsilon \rightarrow 0$, the ignorance region has the form
\begin{align*}
\pi_{Y \mid do(a)}
\in (1-\pi_U) P(Y=1 \mid A=a) + [0, \pi_U].
\end{align*}
Similarly, for cases where $\pi_{U \mid a}$ approaches 1, the data provide no information about $P(Y \mid U = 0, A= a)$, and the ignorance region has width equal to $1-\pi_U$.
This regime appears in Figure~\ref{fig:binary_ignorance} for small and large values of $S(a)$, where the ignorance region widths approach $\pi_U = 0.3$ and $1 - \pi_U = 0.7$, respectively.
As $m$ grows large, the probability that a sampled cause vector $A$ falls in this regime approaches 1.

\subsection{Consistency and Positivity in General}

The above positivity violation occurs in any case where the latent confounder $U$ can be reconstructed consistently from the causes $A$.
We summarize this fact in the following supporting proposition for Theorem~\ref{thm:impossible}.
The central idea of this result is as follows: when $\hat U(A)$ is consistent in the large $m$ limit, the event $U=u$ implies that $A$ takes a value $a$ such that $\hat U(a) = u$.
Thus, for distinct values of the latent variable $U$, the observed causes $A$ must lie in disjoint regions of the cause space, violating positivity.

\begin{proposition}
\label{prop:mutex}
Suppose that Assumption~\ref{assn:multicause} holds, that $P(U)$ is not degenerate, and that there exists a consistent estimator $\hat U(A_m)$ of $U$ as $m$ grows large.
Then positivity is violated as $m$ grows large.
\begin{proof}
Because $P(U)$ is non-degenerate, $U$ takes on multiple values with positive probability.
In this case, we establish that $A_m$ concentrates in different sets of $\mathcal A_m$ depending on the value of $U$.
For each $m$ and each latent variable value $u$ in the support of $U$, define the set
$$E_m(u) = \{a_m : \hat U(a_m) \neq u\}.$$
$E_m(u)$ is the set of cause vector values $a_m \in \mathcal A_m$ that $\hat U(\cdot)$ would map to a value \emph{other} than $u$.
Because $\hat U(A_m)$ is consistent, for each $u$ in the support of $U$, as $m$ grows large,
$$ P(A_m \in E_m(u) \mid U=u) = P(\hat U(A_m) \neq u \mid U = u) \rightarrow 0.$$
Likewise, for any $u' \neq u$ in the support of $U$,
$$ P(A \in E_m(u) \mid U = u') = P(\hat U(A_m) \neq u \mid U = u') \rightarrow 1.$$
Thus, positivity is violated.
\end{proof}
\end{proposition}

\subsection{Proof of Theorem~\ref{thm:impossible}}
We conclude with a proof of Theorem~\ref{thm:impossible}.
\begin{proof}
One of two cases must hold: $P(U \mid A)$ is either degenerate almost everywhere, or not.
In the non-degenerate case, Proposition~\ref{prop:copula} proves non-identification, except in the trivial case.

In the degnerate case, there are again two cases: either $P(U)$ is degenerate, or not.
If $P(U)$ is not degenerate, Proposition~\ref{prop:mutex} shows that the positivity assumption fails, and because $P(Y \mid U, A)$ is nonparametric by assumption, $P(Y \mid U, A)$ is inestimable for some $(u, a) \in \mathcal U \times \mathcal A$, and \eqref{eq:causal integral} is not identified.
If $P(U)$ is degenerate, then the latent variable does not induce any confounding, and $P(Y \mid do(A)) = P(Y \mid A)$.
\end{proof}

\section{DISCUSSION}
\subsection{Practical Implications}
\label{sec:estimation}
In this section, we discuss practical implications of our negative results, then end on a more hopeful note, and suggest some constructive ways forward.

We put the bad news first.
Theorem~\ref{thm:impossible} suggests that one should generally be cautious about drawing causal inferences in the multi-cause setting.
Specifically, the existence of a non-trivial ignorance region can make conclusions highly dependent on modeling choices that in other contexts would be innocuous.
On one hand, flexible, nonparametric models are likely to behave unpredictably when they are used to estimate causal effects in this setting.
In particular, within the ignorance region, point estimates will be driven primarily by choices of regularization.
To demonstrate this, we plot the results of an estimation experiment on the right of Figure~\ref{fig:binary_ignorance}.
In this experiment, we estimate $\pi_{y \mid do(a)}$ for a given vector $a$ from 15,000 datapoints drawn from the binary example's data generating process.
We perform estimation by maximum likelihood, but we add a weak L2 penalty term on the log-odds ratio $\gamma := \log\left(\frac{p_{11 \mid a}}{p_{10 \mid a}}\left / \frac{p_{01 \mid a}}{p_{00 \mid a}}\right.\right)$, which pushes estimates of $p_{11 \mid a}$ to take specific values.
As expected, the penalty determines where in the ignorance region the estimate appears.
This behavior can be unpredictable in complex problems, especially when applying estimation methods that involve tuning parameters or stochastic training.

On the other hand, one may be able to obtain numerically stable estimates of causal effects with parametric assumptions in this setting, e.g., assuming that $E[Y \mid U, A]$ is linear, but our results still suggest caution, unless one has strong prior knowledge that the parametric specification is correct.
Parametric assumptions can induce identification in this setting, but this can be difficult to confirm without formal analysis of the model, as our first counterexample suggests.
In addition, because of the lack of nonparametric identification, if one does obtain identification via parametric assumptions, the parts of these assumptions that determine result of the analysis cannot be checked: the data will be indifferent to different parametric specifications that yield different estimates in the ignorance region.
Finally, uncertainty estimates obtained under parametric specifications may not faithfully represent this sensitivity to functional form.

\subsection{Proxy Variables}
Despite these negative results, there are straightforward modifications to the multi-cause setting that can yield nonparametric identification.
One alternative is to incorporate \emph{proxy variables} (a.k.a. negative controls) into the estimation procedure.
A proxy variable $Z$ is a supplementary variable that is downstream of the latent confounder $U$, but is causally unrelated to either the outcome $Y$ or the cause variables $A$.
Several authors have given sufficient conditions for causal identification when using proxy variables in the presence of unobserved confounding
 \citep[see, e.g.,][] {kuroki2014measurement,Miao_Identifying_2016,shi2018multiply}.
Recently, \citet{louizos2017causal} applied these results to estimate causal effects by calculating \eqref{eq:causal integral} from the posterior distribution of a variational autoencoder model.

Adding proxy variables to an analysis requires collecting a slightly larger dataset, but it is plausible that in many multi-cause settings collecting the appropriate variables would not be onerous.
For example, the sufficient conditions for nonparametric identification in \citet{Miao_Identifying_2016} require the addition of two proxy variables, one conditionally independent of $A$ given $U$, and the other conditionally independent of $Y$ given $U$.
In the multi-cause setting, it is not difficult to find a proxy of the first type; each cause variable $A^{(k)}$ satisfies this condition.
Similarly, it is plausible that proxies of the second type would be readily available in many contexts that could be framed as multi-cause causal inference problems, e.g., GWAS \citep{tran2017implicit} or recommender systems \citep{wang2018deconfounded}.
For example, if the number of causes $m$ is large because $A$ includes causes that span a large set of functions, then it may be plausible to designate some loosely related causes as proxies of this second type.
Likewise, in cases where the latent confounder is shared by two outcomes $Y^{(1)}$ and $Y^{(2)}$ that are conditionally independent given $U$, each outcome can serve as a proxy in estimating the intervention distribution of the other.

To demonstrate that the proxy approach addresses some of the pathologies highlighted here, we repeat the estimation experiment on the right side of Figure~\ref{fig:binary_ignorance}, adding two proxy variables to the data. 
As in Section~\ref{sec:estimation}, we estimate the model parameters by maximum likelihood estimation for varying settings of a weak regularizer.
In this case, the regularizer has negligible effect on the estimates of $\pi_{Y \mid do(a)}$, although the estimates show significant variability.

\subsection{Sensitivity Analysis}
To close, we note that the multi-cause setting could present a fruitful use case for sensitivity analysis as an alternative to point estimation of causal estimands.
In such a case, rather than focus on point-identifying a causal estimand under strong assumptions, it can be useful to relax our requirements, and attempt to identify an ignorance region under weaker assumptions \citep[see, e.g., ][]{manski2009identification}.
The multi-cause setting is particularly amenable to sensitivity analysis, because the assumptions about factorizations of $P(A)$ represent a reasonable middle ground between the strong assumption of no unobserved confounding at one extreme and the weak assumption of unstructured confounding on the other.
For example, although we presented it as a negative result before, Assumption~\ref{assn:multicause} induced ignorance regions obtained in Figure~\ref{fig:binary_ignorance} that are relatively narrow for some values of $a$; for some applications, analyses that explicitly map these sensitivity regions could acceptably informative.

As a bonus, sensitivity analyses can often be applied \emph{post hoc}, so that an investigator is first free to find a best-fitting model for the observed data, and then reason about the causal implications of the model separately \citep{franks2018flexible}.
We leave adaptation of such methods to the multi-cause setting for future work.

\subsubsection*{Acknowledgements}

Thanks to Yixin Wang, Rajesh Ranganath, Dustin Tran, and David Blei for open conversations about their work. Thanks to Alexander Franks, Andrew Miller, Avi Feller, Jasper Snoek, and D. Sculley for their feedback on this project.

\bibliographystyle{chicago}
\bibliography{readcube_export}

\end{document}